# Adversarial Risk and the Dangers of Evaluating Against Weak Attacks

Jonathan Uesato [1]  Brendan O'Donoghue [1]  Aaron van den Oord [1]  Pushmeet Kohli [1]

## Abstract

This paper investigates recently proposed approaches for defending against adversarial examples and evaluating adversarial robustness. We motivate *adversarial risk* as an objective for achieving models robust to worst-case inputs. We then frame commonly used attacks and evaluation metrics as defining a tractable surrogate objective to the true adversarial risk. This suggests that models may optimize this surrogate rather than the true adversarial risk. We formalize this notion as *obscurity to an adversary*, and develop tools and heuristics for identifying obscured models and designing transparent models. We demonstrate that this is a significant problem in practice by repurposing gradient-free optimization techniques into adversarial attacks, which we use to decrease the accuracy of several recently proposed defenses to near zero. Our hope is that our formulations and results will help researchers to develop more powerful defenses.

## 1. Introduction

Deep learning has revolutionized the field of machine learning and led to substantial improvements in many challenging problems such as image understanding (He et al., 2016), speech recognition (Graves et al., 2013), and automatic game playing (Mnih et al., 2015). Despite these remarkable successes, we have seen some intriguing and troubling properties in the behaviour of these models.

Researchers have demonstrated that certain small perturbations to the input can make neural networks produce extremely bad results (Szegedy et al., 2013; Jia and Liang, 2017) For instance, in the case of image classification, imperceptible perturbations can lead to the resulting images being misclassified to completely different object categories with high confidence (Szegedy et al., 2013). These so-called

[1]DeepMind. Correspondence to: Jonathan Uesato <juesato@google.com>.

*adversarial* examples can be computed relatively easily by using optimization methods (referred to as adversarial attacks) to find perturbations that maximize the loss of the network (Goodfellow et al., 2014; Carlini and Wagner, 2017b). Later work showed that this phenomena is not unique to image classification, and appears across different model architectures, as well as in other machine learning algorithms (Papernot et al., 2016; 2017).

The emergence of adversarial examples and the increasing deployment of machine learning models in real world production systems has motivated extensive research on developing models that can defend against adversarial attacks (Warde-Farley and Goodfellow, 2016; Yuan et al., 2017). Using a variety of approaches, these works have shown that neural models can be developed that are robust against commonly used attacks (Guo et al., 2017; Song et al., 2017; Xie et al., 2017; Liao et al., 2017). However, a key question remains unanswered: *Are these models free from any adversarial examples or are they simply robust to current attack methods?*

In this paper, we formalize the intuition that in settings with the potential for catastrophic failures, minimizing expected risk may produce models with very poor worst-case performance. This motivates the study of the adversarial risk as a measure of the model's performance on worst-case inputs. However, the exact adversarial risk is computationally intractable to evaluate, let alone optimize. We thus frame commonly used attacks and adversarial evaluation metrics as defining a tractable surrogate objective to the true adversarial risk. We hypothesize that many defenses achieve robustness through *obscurity*, *i.e.*, the defenses work by exploiting weaknesses of certain attacks and do not eliminate all adversarial inputs.

One of the key contributions of this paper is to experimentally validate the 'security by obscurity' nature of recently proposed defense methods. Specifically, we show that by using a more powerful attack method (one better able to maximize the true adversarial risk), we can dramatically reduce the performance of these defenses.

To summarize, the key contributions of this paper are:

- Formulation of adversarial attacks and defenses as optimizing *surrogates* of the true adversarial risk

- Developing the notion of *obscurity to an adversary* as a tool for reasoning about when models optimize the surrogate without optimizing the true adversarial risk

- Empirical demonstration that many defense approaches achieve security via obscurity, and provide no benefits in true adversarial robustness

- Use of gradient-free optimization techniques to measure when models are obscured to transfer-based and gradient-based attacks (*e.g.*, via gradient masking (Papernot et al., 2017)).

## 2. Related Work

**Adversarial attacks in practice**: Initial work on adversarial examples focused on 'white-box' attacks that had access to the model directly (Goodfellow et al., 2014; Carlini and Wagner, 2017a). Subsequent work also demonstrated that image classifiers can be fooled without direct access to the model, so called 'black-box' or 'oblivious' attacks, (Papernot et al., 2017; 2016), even when using printed images or 3D physical objects (Kurakin et al., 2016b; Brown et al., 2017; Athalye et al., 2018b). While adversarial examples are the most straightforward to study in the visual domain, they have also been demonstrated in other domains such as reinforcement learning (Huang et al., 2017), robotics (Pinto et al., 2017), malware detection (Grosse et al., 2016), and natural language processing (Jia and Liang, 2017).

**Progress on black-box robustness**: While we argue in this work that much of the reported progress in white-box robustness is illusory, there has been progress in defending against black-box attacks, including Ensemble Adversarial Training (Tramèr et al., 2017) and High-level Guided Denoising (Liao et al., 2017). We point interested readers to the NIPS competition on adversarial defenses for a head-to-head comparison of defenses in the black-box threat model, when the adversary has no access to the model and does not know which defense is being used (Kurakin and Goodfellow, 2017).

**Evaluating adversarial robustness**: Previous work has observed the phenomena of *gradient masking* (Papernot et al., 2017; Goodfellow and Papernot, 2017) which refers to the fact that it is easy to intentionally or unintentionally create models without useful gradients either by removing them, making them too small, or adding noise. This can cause gradient-based attacks to fail, which might lead a practitioner to believe a model is robust, when in fact a more sophisticated adversary could generate adversarial inputs for the model. While finalizing this manuscript, Athalye et al. (2018a) published similar conclusions and showed that many adversarial defenses contain components which merely mask gradients. While their attack methods differ, their results similarly underscore the importance of attacking proposed defenses with appropriately designed adversaries.

**Certificates of adversarial robustness**: Our work emphasizes the need for both principled methods of approaching adversarial robustness, and stronger techniques for analyzing neural networks. In this vein, we note the growing community on verification of neural networks (Katz et al., 2017; Bunel et al., 2017; Carlini et al., 2017), which provide techniques to compute provably worst-case adversarial examples. Of particular interest are training procedures which optimize certificates of robustness (Kolter and Wong, 2017; Raghunathan et al., 2018; Sinha et al., 2017) to produce models with provable guarantees. In general, the idea is to design a relaxation of the original adversarial optimization problem, which can be efficiently solved to optimality. The role of training is then to ensure that the optimal value of this relaxed problem is close to the optimal value of the original problem.

**Transparency**: Finally, we note connections to research on transparency in machine learning. In addition to the superficial similarities between techniques for adversarial attacks and feature visualization (Olah et al., 2017), the shared goal of designing models which can be efficiently analyzed suggests the potential for fruitful exchange of ideas (Christiano, 2016). While in the model interpretability community, transparency is often measured qualitatively from the perspective of a human (Kim, 2015; Krakovna and Doshi-Velez, 2016; Ross and Doshi-Velez, 2018), we note that obscurity as we define it provides one quantitative measure of the degree to which models can be easily analyzed by other algorithms.

## 3. Adversarial Risk and Obscurity

### 3.1. Adversarial Risk and Worst-Case Guarantees

We formalize the ideas in this paper in the context of supervised learning, although they naturally extend to the reinforcement learning setting. We seek parameters $\theta$ of a model $m_\theta$ that minimize the loss $\ell$ on inputs $x$ and labels $y$ sampled from the data distribution $D$. Typically, the objective is to find parameters $\theta$ which minimize the *expected* risk:

$$\mathop{\mathbb{E}}_{(x,y)\sim D} \ell(m_\theta(x), y). \qquad (1)$$

In practice, this is done by optimizing the empirical risk on a finite training set, and estimating the expected risk with the average loss over a held-out test set. We note two related issues. First, even models with low expected risk may perform arbitrarily poorly on any given data point. This may be problematic for deployment of machine learning systems in very high-stakes situations, where even a single catastrophic failure is not acceptable. Second, if the loss function is unbounded, the variance of the estimator may be arbitrarily large (Brownlees et al., 2015; Audibert et al., 2011). In these cases, we might desire models which also have small *worst-case* risk:

$$\sup_{(x,y) \in \text{supp}(D)} \ell(m_\theta(x), y), \quad (2)$$

where $\text{supp}(D)$ denotes the support of $D$. The difficulty arises from the fact that the supremum may not be easy to compute. First, while the expectation in (1) can be estimated by samples from the data distribution, naively computing the supremum in (2) requires exhaustively searching the input space, which could be exponentially expensive in the dimension of $x$. Second, performing a search would require already knowing the shape of the distribution's support, but the mapping from $x$ to $y$ is exactly what we seek to learn.

We view the local *adversarial risk*, denoted $L$, as a proxy for the worst-case risk:

$$L(\theta) = \mathbb{E}_{(x,y) \sim D} \left[ \sup_{x' \in N_\epsilon(x)} \ell(m_\theta(x'), y) \right], \quad (3)$$

where the neighborhood $N_\epsilon(x)$ denotes the set of points in $\text{supp}(D)$ within a fixed distance $\epsilon > 0$ of $x$, as measured by some metric. By replacing the global search for the highest possible loss over the entire input space with a local search over $N_\epsilon(x)$, the adversarial risk formulation allows us to use off-the-shelf optimization algorithms, such as projected gradient descent (Madry et al., 2017; Kurakin et al., 2016a). Further, sufficiently constraining $\epsilon$ such that the true label $y$ does not change within this region allows us to search for approximately worst-case examples, without knowing the exact mapping from $x$ to $y$.

While the local adversarial risk only provides us a certificate for a model at a fixed set of points, as opposed to worst-case risk which provides a certificate for a model on the whole, it allows us to study the challenges of obtaining worst-case guarantees in a simpler setting, and is also a necessary sub-problem, as the adversarial risk is a lower bound on the worst-case risk.

### 3.2. Obscurity with Respect to an Adversary

The true adversarial risk $L$ is still difficult to compute exactly due to the inner supremum. This is typically approximated by using a particular choice of adversary $f$, which generates an adversarial example in $N_\epsilon(x)$ (for example, by using $k$ steps of projected gradient descent). We use $\hat{L}$ to denote this surrogate adversarial risk, measured against a particular adversary:

$$\hat{L}(\theta, f) = \mathbb{E}_{(x,y) \sim D} \ell(m_\theta(f(\theta, x, y)), y). \quad (4)$$

We frequently refer to this as the (adversarial) evaluation metric, measured against a particular adversary, because it is the quantity reported in papers. However, without additional information, good performance on the evaluation metric need not imply high adversarial robustness. Unlike in standard supervised learning, where the test loss gives an unbiased estimator of the expected risk, the surrogate adversarial risk is biased downwards from the true adversarial risk. As we will show later, the gap between these two terms can be large.

We note two potential explanations for models with high evaluation performance: models which learn a robust decision boundary, and thus admit few adversarial examples, and models which achieve *security via obscurity*. That is, while their decision boundaries admit adversarial examples, the model makes adversarial examples difficult to identify efficiently (Papernot et al., 2017). In the case of gradient-based adversaries, a model may exploit gradient masking to bypass the specific adversary, without achieving true adversarial robustness (Tramèr et al., 2017; Goodfellow and Papernot, 2017).

We refer to the difference between these terms, as the model's *obscurity to the adversary $f$*, i.e.,

$$\text{Obscurity}(\theta, f) = L(\theta) - \hat{L}(\theta, f). \quad (5)$$

Intuitively, the model has high obscurity with respect to $f$ if the model admits many adversarial examples which $f$ fails to find. We will say a model is transparent if we have a computationally efficient adversary $f$, against which the model has low obscurity.

We note three reasons to actively seek transparent models when designing adversarial defenses. First, without transparent models, we lack the tools to compute the true adversarial risk. From a scientific perspective, this makes it difficult to assess whether high evaluation scores reflect strong defenses or weak attacks. Second, transparency is necessary if we are to know in advance of deployment whether a model may fail catastrophically or not. In some cases, it may be important to know before deploying a machine learning model whether that model can fail in an unlikely but catastrophic manner. Finally, as we discuss in Section 6.4, to the extent that adversarial training may be used to train highly robust models (Madry et al., 2017), the ability to efficiently identify existing adversarial examples is a practical requirement. In other words, if adversarial training works well, efficient adversarial testing techniques become key to achieving low adversarial risk.

## 4. Attack Strategies

We seek to answer the empirical question: *what is the obscurity of a model to adversaries commonly used in practice?* In general, computing the inner maximization in (3) is not tractable. However, *any* adversary may be used to compute lower bounds on the true adversarial risk, which also provides a lower bound on the obscurity with respect to a fixed adversary.

In this section, we describe both gradient-based and gradient-free optimization strategies, which we use to demonstrate that many seemingly robust defenses in fact admit adversarial examples. We contrast these with transfer-based attacks, which are frequently used for evaluation, but are more difficult to interpret due to the subtleties in designing a surrogate model similar to the original model.

### 4.1. Optimization-based attacks

We frame finding an (untargeted, white-box) adversarial perturbation as a constrained optimization problem with a margin-based loss, which is negative when an image is misclassified (Carlini and Wagner, 2017b), and we use the $\ell_\infty$ norm to determine the neighborhood around the input point $x_0$ with label $y_0$:

$$\begin{aligned} \min_x \quad & m_\theta(x)_{y_0} - \max_{j \neq y_0} m_\theta(x)_j \\ \text{s.t.} \quad & \|x - x_0\|_\infty < \epsilon, \end{aligned} \quad (6)$$

where $m_\theta(x)_j$ denotes the output logit for the class $j$. This loss, which we denote $J_\theta^{\text{adv}}(x)$, has been shown to be easier to optimize than the cross-entropy loss because logits behave more 'linearly' than the output probabilities (Carlini and Wagner, 2017b).

**Gradient-based optimization**: When possible, we use projected gradient descent (PGD) to solve this optimization problem (Kurakin et al., 2016b). In the most basic version, we perform the following update for each iteration:

$$x^+ = \Pi_{N_\epsilon(x_0)}(x + \alpha \nabla_x J_\theta^{\text{adv}}(x)) \quad (7)$$

where $\Pi_{N_\epsilon(x_0)}$ denotes Euclidean projection onto the set $N_\epsilon(x_0)$, and where $\alpha > 0$ is a step-size. In practice, we replace the vanilla gradient update with Adam (Kingma and Ba, 2014), which tends to converge faster for this problem. We additionally initialize $x$ to a random perturbation within $N_\epsilon(x_0)$, in case of poor local minima along the data manifold due to gradient masking (Tramèr et al., 2017).

**Gradient-free optimization**: In cases where we cannot take analytic gradients, or where they are not useful, we can approximate them with finite difference estimates in random directions. In this work, we use SPSA (Spall, 1992), as described in Algorithm 1, which is well-suited for high-dimensional optimization problems, even in the case of noisy objectives (Spall, 1992).

Maryak and Chin (2001) show that the stochasticity introduced by sampling perturbations allows SPSA to converge to the global minimum, under appropriate conditions. While their assumptions to not hold exactly in our setting, they point to a qualitative difference between analytic gradients, which provide the optimal update direction based on an infinitesimally small region, and gradient-free methods, which measure changes resulting from finite perturbations.

**Algorithm 1** SPSA adversarial attack

**Input:** function to minimize $f$, initial image $x_0 \in \mathbb{R}^D$, perturbation size $\delta$, step size $\alpha > 0$, batch size $n$
**for** $t = 0$ **to** $T - 1$ **do**
    Sample $v_1, \ldots, v_n \sim \{1, -1\}^D$
    Define $v_i^{-1} = [v_{i,1}^{-1}, \ldots, v_{i,D}^{-1}]$
    Calculate $g_i = (f(x_t + \delta v_i) - f(x - \delta v_i))v_i^{-1}/(2\delta)$
    Set $x'_t = x_t - \alpha(1/n)\sum_{i=1}^n g_i$
    Project $x_{t+1} = \arg\min_{x \in N_\epsilon(x_0)} \|x'_t - x_0\|$
**end for**

Note that we average gradient estimates to allow for an efficient GPU implementation, and because we observed this to result in faster convergence. Additionally, although previous implementations of SPSA for neural networks tend to sample $v$ from Gaussian distributions (Salimans et al., 2017a; Ilyas et al., 2017) as is done in natural evolution strategies (Wierstra et al., 2008), we sample $v$ from a Rademacher distribution (*i.e.*, Bernoulli $\pm 1$), as SPSA requires $v^{-1}$ to have finite first and second moments (Spall, 1992).

In initial experiments, we also tried natural evolutionary strategies (Wierstra et al., 2008; Ilyas et al., 2017) and zero-order optimization (Chen et al., 2017), two other gradient-free optimization techniques. Comparisons are included in Appendix A. We found SPSA to reliably produce adversarial examples, but we believe the specific choice of gradient-free optimizer is relatively unimportant. In particular, any adversary provides a valid lower bound on the true adversarial risk.

### 4.2. Transfer-based attacks

In cases where the adversary lacks direct access to the model being attacked, a useful attack strategy is to *transfer* adversarial examples from a known model $m'$ to the unknown model $m$. This approach has been effective against undefended classifiers both when the surrogate model is trained to mimic the unknown model (Papernot et al., 2017) and when the surrogate is a generic image classifier (Szegedy et al., 2013). However, while transfer-based methods provide a practical attack vector, robustness to any given transfer-based attack may not imply robustness in general, since the success of the attack is highly dependent on the similarity between $m'$ and $m$.

One particularly relevant case is defenses which apply a (possibly non-differentiable) transformation $t$ before classifying the image. A common evaluation strategy is then to transfer adversarial examples from the original model $m$ to the defended model $m \circ t$. In this case, the defended model may admit adversarial examples, but appear to be adversarially robust, provided its weaknesses are sufficiently different from those of the original model $m$.

| Dataset | Defense strategy | Original Evaluation | Adversarial Accuracy Bound | Obscurity Bound |
|---|---|---|---|---|
| CIFAR-10 ($\epsilon = 8$) | PixelDefend (Song et al., 2017) | 75% | <10% | >65% |
| | Adversarial Training (Madry et al., 2017) | 47% | <47% | >0% |
| ImageNet ($\epsilon = 2$) | Non-differentiability (Guo et al., 2017) | 15% | 0% | 15% |
| | Stochasticity (Xie et al., 2017) | 36% | <1% | >35% |
| | High-level Guided Denoiser (Liao et al., 2017) | 75% | 0% | 75% |

Table 1: **Summary of results.** We find that while most proposed defenses confer significant robustness to the adversaries they are evaluated against in the original papers, they remain highly vulnerable to adversarial examples. Perturbation sizes $\epsilon$ are relative to images with pixel intensities in $[0, 255]$. Note that our numbers in the left column differ slightly from those in the original papers due to standardizing evaluation conditions across defenses. We explain the sources of these differences, compared to the pure replication, in the corresponding subsections.

## 5. Reasoning about Obscurity

A key point is that while neither obscurity nor adversarial risk are directly computable, we can reason a priori about obscurity much more easily than adversarial risk for new models. From the viewpoint of obscurity, the relevant question is not "does the model admit any adversarial inputs?" but rather: *"if the model were to admit adversarial inputs, would the adversarial evaluation detect these inputs?"* Thus, while the adversarial robustness of a particular defense ultimately relies on empirical evaluation, obscurity can guide algorithm design towards approaches which will not be obscured to adversaries, which confers significant advantages as discussed above in Section 3.2. We briefly outline several heuristics here, which are further developed in context throughout the experimental section:

*Will the surrogate objective converge to the true adversarial objective?* Many evaluations with transfer-based adversaries fail this test, since even in the limit, an adversarial example for the surrogate classifier $m'$ might not fool the original classifier $m$, unless $m'$ and $m$ are guaranteed to be similar. Of course, in some cases adversarial examples may transfer, as in Papernot et al. (2017; 2016), and indeed if the surrogate is optimized to match the original model, such attacks can be highly effective. But on its own, merely sharing some parameters does not guarantee adversarial examples will transfer from the surrogate to the defense model. In Section 6.2.3, we provide an example of a model which suffers almost no change in accuracy against a transfer-based attack, despite having $0\%$ true adversarial accuracy.

*Is the true objective well-behaved?* (Goodfellow et al., 2014) argue that gradient-based search techniques for adversarial examples work for largely the same reasons that gradient-based optimization works for training deep networks. From the view of obscurity and robustness to worst-case adversaries, this is an advantage, not a disadvantage. This means that standard issues arising in neural network training, such as vanishing gradients (Sec. 6.2.2) and discrete or highly nonlinear operations (Sec. 6.1), also present difficulties for gradient-based adversaries.

Conclusively demonstrating that a defense has high obscurity ultimately does require evaluating against a stronger adversary, as we do experimentally. However, heuristic reasoning can often approximately predict obscurity of a model, thus allowing focus on approaches with low obscurity. We validate these heuristics experimentally by studying several defenses we hypothesized to have high obscurity to their evaluation adversary, and one we did not. We show that in all defenses hypothesized to have high obscurity, the evaluation metric does not capture the true adversarial risk. In these cases, we show that the true adversarial accuracy is near zero in spite of high surrogate adversarial accuracy.

## 6. Experiments

In this section we empirically study several broad categories of proposed defense strategies on the CIFAR-10 (Krizhevsky and Hinton, 2009) and ImageNet (Deng et al., 2009) datasets. To follow standard practice, we assume the $0-1$ loss and report accuracy, i.e. $1 - \hat{L}(\theta, f)$. [1] We selected examples to illustrate pros and cons of each approach and we attempted to use the strongest possible defense within each category where possible (for example, the two highest-scoring submissions to the NIPS Adversarial Defenses Competition).

We note that many published works combine elements of multiple strategies, such as adding defensive components *ex post* to an adversarially trained network. However, we analyze defense strategies separately, both to isolate the effect of the core idea, and because combining defenses does not affect our underlying claim: that obscured defenses create a gap between the evaluation metric and the true adversarial risk.

---

[1] However, as previously noted, adversarial robustness is most salient in settings with heavy-tailed losses, where the task requires models which avoid catastrophic failures.

## 6.1. Non-differentiability

**Original evaluation**: Guo et al. (2017) argue that non-differentiability combined with stochasticity can provide a useful defense against adversarial examples. They propose several variations on the core idea of applying an image transformation to remove adversarial artifacts before classifying the image with a pretrained classifier. Although JPEG encoding is among the weaker proposed defenses we include it here as a particularly clear example of how common adversaries may fail to measure the true adversarial risk, and note that similar arguments apply to the other proposed defenses.

Guo et al. (2017) evaluate their defenses against a variety of transfer-based attacks, by optimizing an adversarial image to fool the original undefended model, and then classifying these images with the defended model. We verify that JPEG compression, with JPEG quality 75, provides some defense against transfer-based attacks (referred to as "gray-box"), by using PGD to generate high-confidence adversarial examples, against which the defense achieves 15% accuracy on ImageNet at $\epsilon = 2$, compared to 0% for the undefended model.

**Demonstration of obscurity**: We argue that this defense is obscured to both gradient-based and transfer-based adversaries, and that neither attack should be expected to identify near worst-case adversarial attacks, even against a non-adversarially robust model.

First, standard gradient-based optimizers cannot be used when the model is non-differentiable. However, as previously noted (Goodfellow and Papernot, 2017), any model can trivially be made non-differentiable without removing adversarial examples, for example by decreasing the softmax entropy or converting the softmax to a hard maximum.

Second, as discussed in Section 5, merely sharing parameters used in the original model (gray-box attack) is insufficient to guarantee adversarial examples will transfer from the surrogate to the original model, if the surrogate model uses different preprocessing.

To verify these arguments, we attack the defended model, $m \circ t$ where $t$ represents JPEG compression, using the SPSA adversary. By using a gradient-free method, we are able to attack the end-to-end model, despite the lack of an analytic gradient. Using this attack, we reduce the adversarial accuracy of the model to 0% on ImageNet with $\epsilon = 2$.

## 6.2. Leveraging generative models

### 6.2.1. Detecting adversarial examples

**Original evaluation**: Song et al. (2017) argue that "adversarial examples mainly lie in the low probability regions of the training distribution, regardless of attack types and targeted models." They use a PixelCNN as a density model (van den Oord et al., 2016) to flag inputs with likelihood below some threshold as potentially adversarial. While this is not a method to classify adversarial examples per se, we include it as an illustrative example, since many similar proposals to leverage generative models for adversarial examples appeal to an intuition that density models should flag adversarial examples far from the natural data manifold (Meng and Chen, 2017).

For our experiments, we use a pretrained PixelCNN model, which assigns a likelihood of 3.02 bits/dim on the CIFAR-10 test set. We verify their result that adversarial examples generated by an adversary oblivious to the detection model can be detected with likelihood thresholding. Figure 1 shows that adversarial examples generated with PGD on CIFAR-10 at $\epsilon = 8$ receive significantly lower likelihoods from the PixelCNN, compared to natural images. Specifically, thresholding at 4.68 bits/dim achieves a 1% false postive rate while detecting 99.7% of adversarial examples.

**Demonstration of obscurity**: If the adversary is aware of the defense, we can construct non-detectable adversarial examples by simultaneously optimizing for both misclassification and likelihood under the density model. Formally, letting $J_\theta^{\text{adv}}(x)$ denote the objective in the original adversarial optimization in Eq. 6, and $l_{\text{CNN}}(x)$ denote the negative log-likelihood of $x$ according to the PixelCNN model, we modify the objective in Eq. 6 with $J_\theta^{\text{adv}}(x) + \lambda l_{\text{CNN}}(x)$, and optimize with PGD as usual. This allows us to find adversarial examples which receive *higher* likelihood than their natural counterparts, as summarized in Figure 1. Using the same threshold, to achieve a 1% false positive rate, the detector now detects 0% of the adversarial examples.

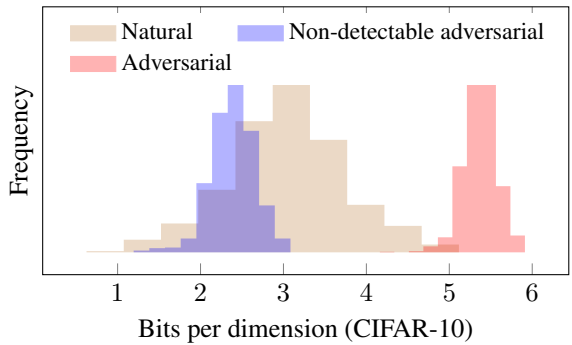

Figure 1: **Detecting adversarial examples with a generative model**: While typical adversarial examples lie in low probability regions of the input space, an adversary with access to the density model can construct non-detectable adversarial examples, which are both misclassified and assigned higher likelihood than than their natural counterparts.

### 6.2.2. PURIFICATION VIA LIKELIHOOD OPTIMIZATION

**Original evaluation**: The main defense proposed in Song et al. (2017) is the *PixelDefend* procedure, which uses a PixelCNN to purify images towards the distribution observed in training. Concretely, the PixelDefend algorithm involves maximizing the likelihood $l_{\text{CNN}}$, subject to a neighborhood constraint, similar to eq. 6, but using a different objective. The constraint can be modified to balance clean accuracy with adversarial accuracy; we use $\epsilon_{\text{defend}} = 16$, which achieved best results in the original paper.

Song et al. (2017) reported that optimizing likelihood with gradient descent resulted in lower rather than higher likelihoods, and instead opted for an iterative greedy algorithm, which iterates over the pixels and sets each to the value within the feasible set which maximizes the likelihood of the new image. However, we found that when using a continuous, rather than discrete, parameterization over pixel likelihoods, gradient descent was able to optimize the likelihood. [2] We used the gradient-based version of PixelDefend since it avoids the need to evaluate the PixelCNN model once per pixel.

Song et al. (2017) consider two types of attacks: a gradient-based attack against the end-to-end model, and a transfer-based attack, using the undefended network as the surrogate model. We verify that PixelDefend is able to successfully defend against the strongest attack, transfer-based FGSM, achieving 75% accuracy at $\epsilon = 8$.

**Demonstration of obscurity**: As stated in Song et al. (2017), an end-to-end gradient-based attack against the PixelDefend model is impractical, due to vanishing gradients when backpropagating through the unrolling of the optimization process. We avoid this issue by using SPSA and directly attacking the end-to-end model, which reduces the accuracy of PixelDefend to 10% on CIFAR-10 at $\epsilon = 8$. While we acknowledge that exact values may differ due to differences in implementation, we believe these numbers neverthelesss reflect the fact that the lengthy optimization process obscures the model to the gradient-based and transfer-based adversaries in the original paper.

We note that finding adversarial examples for the PixelDefend with SPSA is computationally expensive, due to both the optimization step in the PixelDefend inference procedure, and the sample inefficiency of gradient free optimization. In this sense, the computational cost of the PixelDefend model provides a form of obscurity, as it raises the concern that additional adversarial examples could be found simply by running a more expensive optimization.

---

[2] Precisely, the likelihood over pixels is parameterized as a mixture of logistics as in PixelCNN++ (Salimans et al., 2017b), rather than a 256-way softmax.

### 6.2.3. PURIFICATION WITH AUTOENCODERS

**Original evaluation**: Another approach to bringing adversarial examples closer to the data manifold is to train a neural autoencoder to predict purified images from adversarial images. In Liao et al. (2017), the autoencoder can either be trained to produce images which are similar in pixel space, by using the distance between the clean image and purified adversarial image as the loss, or in feature space, by using the distance between the classifier's logits on the clean and purified images as the loss. Using the latter approach, which they name the *high-level guided denoiser*, they won the NIPS Adversarial Examples competition, demonstrating strong performance against over 100 oblivious black-box adversaries (Kurakin and Goodfellow, 2017).

In Liao et al. (2017), the 'white-box' evaluation is a transfer-based attack, using the undefended model as a surrogate. In our replication, the defended model achieves 75% accuracy on ImageNet at $\epsilon = 2$.

**Demonstration of obscurity**: Since the autoencoder $t(x)$ is a fully differentiable function from images to images, we can simply view the end-to-end model $m \circ t$ as a single neural network, and attack the entire model with PGD. Using this strategy, we decrease the model's accuracy to 0% on ImageNet at $\epsilon = 2$. We note that our results echo those in Gu and Rigazio (2014), which observed on MNIST that while preprocessing images with an autoencoder provided defense against an oblivious adversary, the end-to-end model was actually more vulnerable to adversarial examples from a white-box adversary.

This example supports the point in Section 5 that, unless the surrogate model in transfer-based attacks is somehow *optimized* to produce examples transferable to the original model, we should not expect the attack to necessarily succeed even against non-adversarially robust defenses. In this case, the model suffers almost no performance drop from a transfer-based attack, despite having 0% accuracy against an white-box optimization-based adversary.

### 6.3. Stochasticity and Ensembling

**Original evaluation**: Xie et al. (2017) propose using randomized models, so that an adversary is unable to know which model to optimize against. Rather than training an entire ensemble of models from scratch, they propose to augment a single model with random amounts of padding and resizing, based on the observation that small resizing operations can remove many adversarial effects. Combined with Ensemble Adversarial Training (Tramèr et al., 2017), this model ranked second in the NIPS Adversarial Defenses competition, demonstrating strong performance against oblivious black-box adversaries.

The strongest adversary used for evaluation is the authors'

proposed 'ensemble-pattern attack.' For this attack, they construct a deterministic surrogate model, by selecting $k = 21$ sub-models, each consisting of a fixed padding-resize template which is applied before classifying the image, and averaging the predictions of the sub-models. We verified that the stochastic model has moderate robustness against this attack, with 36% accuracy on ImageNet at $\epsilon = 2$. [3]

**Demonstration of obscurity**: As before, we note even a non-adversarially robust model may demonstrate anything between zero and clean accuracy against transfer-based attacks, depending on the similarity of the surrogate model. While the authors partially address this by using an ensemble as the surrogate, which improves transferability, it is still possible that the obtained adversarial example could overfit to the specific sub-models.

To attack the model, we perform stochastic gradient descent directly on the expectation of the objective. Formally, we replace the original objective $J_\theta^{\text{adv}}(x)$ in Eq. 6 with $\mathbb{E}[J_\theta^{\text{adv}}(x)]$, where the expectation is over randomness internal to the model. Using stochastic gradient descent we can maximize this term, the expected margin of misclassification, by sampling from the distribution of random paddings and resizings, and performing a gradient step evaluated for that specific sub-model (Athalye et al., 2018b). Using this attack, we decrease the accuracy of the stochastic model to 1% on ImageNet at $\epsilon = 2$.

### 6.4. Adversarial training

**Original defense**: In adversarial training (Goodfellow et al., 2014), the network is trained to minimize the adversarial risk by performing stochastic gradient descent at the (approximately) maximally adversarial perturbations of the training data. This is done by generating adversarial examples at each training iteration, and training on the adversarial examples, rather than the natural data. Following Madry et al. (2017) our adversarially trained network achieves 47% accuracy on CIFAR-10 against a gradient-based adversary with $\epsilon = 8$.

**Discussion of obscurity**: Unlike the previously discussed defenses, we did not (*a priori*) suspect adversarially trained networks to be obscured to the evaluation adversary. However, we wanted to investigate further: while adversarial training does not explicitly introduce any non-differentiable components, the network could still learn to mask gradients through the training procedure. For example, Tramèr et al. (2017) demonstrate that adversarial training with a single-step adversary results in gradient masking, unless the adversary initializes to random perturbations.

---

[3] Our numbers differ from the numbers reported for white-box accuracy in the original paper because they evaluate against a subset of the test set which every member of a fixed ensemble classifies correctly, whereas we simply pick 1000 images at random.

We use gradient-free optimization as a tool to check whether a more subtle version of this phenomena occurs even when training with stronger adversaries – intuitively, if the loss surface is highly nonconvex, finite perturbations may be more informative than infinitesimal ones. However, we found that the model achieved 51% accuracy against SPSA at $\epsilon = 8$, and only in $0.2\%$ of images did SPSA succeed when PGD did not. Although far from conclusive, this adds to the picture of PGD as a near-optimal adversary for differentiable image classifiers: alongside evidence that random restarts of PGD converge to similar loss values (Madry et al., 2017) and that PGD is a near-optimal adversary in small neural networks (Carlini et al., 2017). In Appendix C, we provide additional experiments supporting this view.

As one of the few techniques which currently appear to improve the true adversarial risk, adversarial training underscores the importance of transparent models with low obscurity to an efficient adversary. For example, current versions of adversarial training would not be viable with non-differentiable models, because gradient-free optimizers, such as SPSA, are orders of magnitude less efficient than their gradient-based counterparts.

## 7. Conclusion

We believe that as machine learning continues to be deployed in increasingly high-stakes applications, assessing and improving the worst-case behavior of models will become increasingly important. While the ultimate goal is models with low worst-case risk (as well as low expected risk), we view adversarial risk as a simpler proxy objective which allows us to study similar challenges of obtaining worst-case guarantees in machine learning models.

We thus frame common evaluation procedures as defining a tractable surrogate to the true adversarial risk. In this view, the evaluation metric with any fixed adversary is only relevant to the extent that we would expect the adversary to find adversarial examples if they exist.

We believe that the framework of obscurity to adversaries offers a useful, if rough, sieve for designing adversarial defenses, and develop several heuristics for designing transparent defenses. In our experimental validation, we confirmed that in cases where we suspected obscurity to the evaluation adversary, we were able to devise attacks which resulted in dramatically lower adversarial accuracy.

Recent work has seen significant progress in robustness to oblivious adversaries (Liao et al., 2017; Tramèr et al., 2017; Xie et al., 2017), largely driven by standardized blackbox evaluation procedures (Kurakin and Goodfellow, 2017). Similarly, we hope that rigorous evaluation against strong adversaries and better understanding of obscurity will drive progress towards models free of adversarial examples.

# Acknowledgments

We would like to thank Wojciech Czarnecki, Krishnamurthy Dvijotham, Thore Graepel, Koray Kavukcuoglu, Heinrich Kuttler, Ananya Kumar, Janos Kramar, and Valentin Dalibard for their feedback in the preparation of this manuscript, and Fangzhou Liao and Liang Ming for useful discussions about their paper. We are also grateful to Paul Christiano and Arka Pal for early discussion of these ideas, as well as many others on the DeepMind team for providing insightful discussions and support.

## A. Comparison of Gradient-free Attacks

In our initial experiments, we evaluated several different gradient-free attacks, including SPSA, natural evolutionary strategies (NES) (Wierstra et al., 2008; Ilyas et al., 2017), and zero-order optimization (ZOO) (Chen et al., 2017). For ZOO, we used ZOO-ADAM, which uses coordinate descent with coordinate-wise Adam updates (Kingma and Ba, 2014), without the importance sampling or feature reduction tricks used on ImageNet.

All our initial experiments on ImageNet were run against a ResNet-50 model, which achieves 75% accuracy, with 100 randomly selected test images. All experiments on CIFAR-10 used a VGG-like fully convolutional model, which achieves 94.5% accuracy, with 1000 randomly selected test images.

**Overall attack success rate**: On CIFAR-10, all attacks drive accuracy to 0%, using less than 32768 model evaluations, and often much fewer, *e.g.* SPSA decreases accuracy to less than 5% after 2048 model evaluations. Attacks on ImageNet tended to require substantially more model evaluations to drive accuracy to 0%, likely due to the high dimensionality of the input images. Results on ImageNet are summarized below in Table 2.

| Batch size | SPSA | NES | ZOO |
| --- | --- | --- | --- |
| 256 | 29% | 30% | 72% |
| 512 | 21% | 22% | 69% |
| 2048 | 2% | 2% | 43% |
| 8192 | 0% | 0% | 10% |

Table 2: **Attack success rate with increasing computation** We show the accuracy of ResNet-50 against each adversary, with varying amounts of computation. Each attack is run for a maximum of 300 iterations, but with varying batch sizes. In other words, we vary the number of finite difference estimates used before applying each gradient estimate. With sufficient computation, both NES and SPSA decrease accuracy to 0% on ImageNet with $\epsilon = 2$.

**Convergence speed**: Figure 2 provides a different view on the same data and shows the amount of computation before the adversary finds a misclassified image.

For our main experiments, we use a fixed large batch size because our aim is to produce models robust to the strongest possible attacks. In particular, the hyperparameters are selected to reliably generate adversarial examples, rather than to be query-efficient, but in situations where query efficiency is a factor, the same algorithms can be tuned by, for example, reducing the batch size.

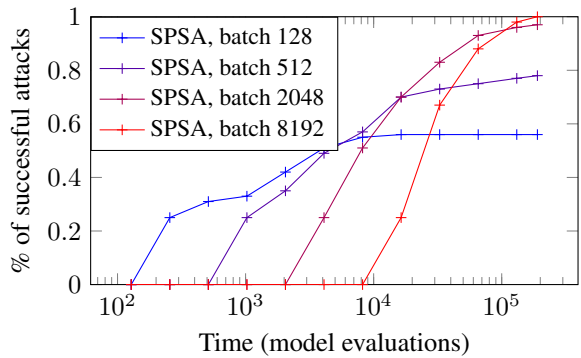

Figure 2: **Model evaluations before finding misclassified adversarial example**: Although attacks with larger batch sizes require longer before beginning to identify adversarial examples, over time, they more reliably identify adversarial examples. For fair comparison, attacks with smaller batch sizes uses more iterations so that the number of model evaluations is constant across attacks. We use a log scale axis since most attacks succeed significantly sooner than the maximum number of iterations.

## B. Hyperparameters

We evaluated all attacks and defenses on 1000 images randomly sampled from either the CIFAR-10 or the ImageNet test set, with the exception of PixelDefend which we could only evaluate on 100 images due to a limited computational budget. We ran attacks for a maximum of 100 iterations, and stopped when the margin-based objective in Eq. (6) was less than $-5.0$. In many cases, the attack completes fairly quickly, in fewer than ten iterations. For all of our optimization-based attacks, we use random initializations by sampling a perturbation from the $\ell_\infty$ ball and projecting back onto the interval $[0, 255]$ (to ensure the resulting image is valid). Hyperparameters are provided below. We have additionally made our implementation avaliable through the open-source library *cleverhans* (Papernot et al., 2018).

| | |
| --- | --- |
| **Perturbation size $\delta$** | 0.01 |
| **Adam learning rate** | 0.01 |
| **Maximum iterations** | 100 |
| **Batch size** | 8192 |

Table 3: **Hyperparameters for SPSA attack**

## C. Discussion of Adversarial Training

As discussed in the paper, gradient-free adversaries allow us to investigate the degree of gradient masking in adversarially trained networks. Gradients estimated based on finite perturbations, rather than infinitesimal ones, may result in qualitatively different behavior (for example, in the case of high-frequency oscillations in the loss surface). Thus, SPSA may, in some cases, converge to better minima than techniques which purely use analytic gradients.

Section 6.4 verified that the expected surrogate adversarial risk (averaged over the test set) cannot be decreased with SPSA. We also investigated whether SPSA could identify better minima than projected gradient descent (PGD) for certain data points. We found that SPSA and PGD found similarly adversarial perturbations for almost all images, which provides further evidence that, in adversarially trained networks, perturbations found by PGD may be nearly worst-case possible perturbations, in which case the surrogate adversarial risk measured against a PGD adversary will be close to the true adversarial risk. These results are summarized in Figure 3.

## D. Additional Experiments

In Figure 4, we include complete experimental results evaluating each of the (non-adversarial training) defenses across a range of perturbation sizes for a number of different attack methods. Although all models shows significant robustness against the original evaluation adversaries, the accuracy of all models falls to near zero when evaluated against stronger attacks.

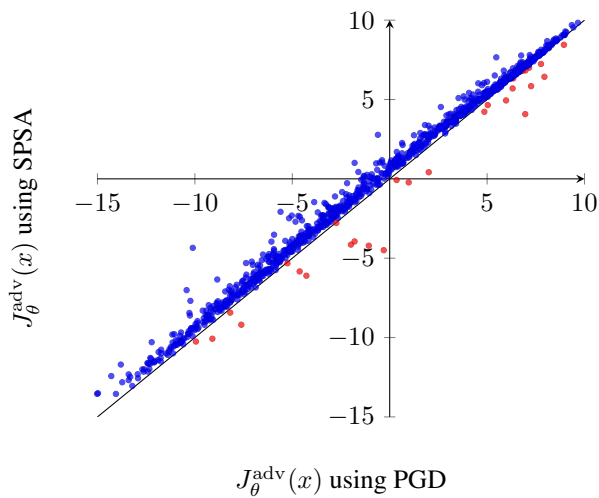

Figure 3: **Analysis of gradient-free masking in adversarially trained networks**: We compare the final values of the margin-based objective across different images, after using projected gradient descent or SPSA. Each point represents a single image, and is misclassified when $J_\theta^{\text{adv}}(x) < 0$. Points close to the line $y = x$ indicate that SPSA and PGD identified similarly adversarial perturbations. Points below the line, shown in red, indicate those for which SPSA identified stronger adversarial perturbation than PGD. Overall, SPSA and PGD identify comparably adversarial perturbations, and there are few points where SPSA identifies significantly stronger adversarial perturbations than PGD.

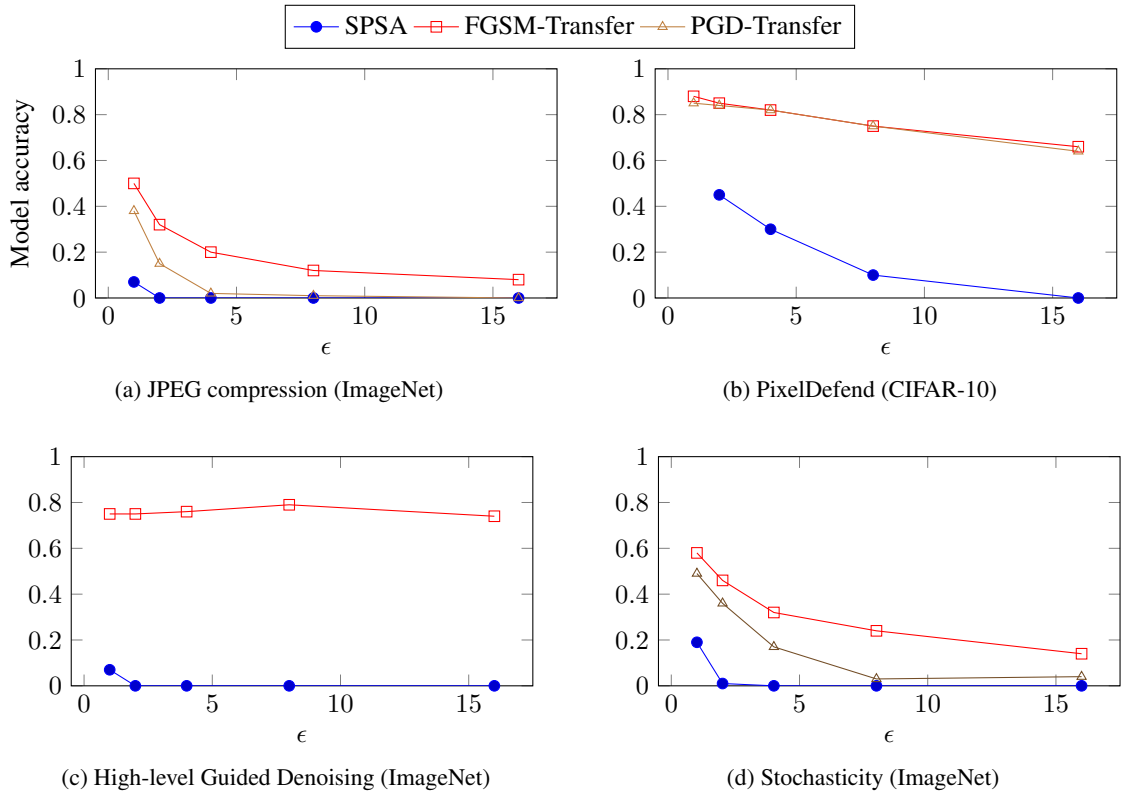

(a) JPEG compression (ImageNet)

(b) PixelDefend (CIFAR-10)

(c) High-level Guided Denoising (ImageNet)

(d) Stochasticity (ImageNet)

Figure 4: Models can be obscured to gradient-based and transfer-based attacks by adding effectively non-differentiable operations and using purification to change the model's decision boundaries compared to the original model. However, this does not remove all adversarial examples – using stronger attacks, we can reduce the accuracy of all defenses to near zero.